\documentclass[11pt]{article}

\usepackage[final]{acl}

\usepackage{times}
\usepackage{latexsym}

\usepackage[T1]{fontenc}

\usepackage[utf8]{inputenc}

\usepackage{microtype}

\usepackage{inconsolata}

\usepackage{graphicx}

\usepackage{booktabs}
\usepackage{amsmath}
\usepackage{amssymb}
\usepackage{multirow}
\usepackage{algorithm}
\usepackage{algpseudocode}
\usepackage{xcolor}

\title{Beyond Alignment: Value Diversity as a Collective Property in Multicultural Agent Systems}

\author{
Shaoyang Xu$^{1}$ \quad
Jingshen Zhang$^{1}$ \quad
Long P. Hoang$^{1}$ \quad
Jinyuan Li$^{2}$ \quad
Wenxuan Zhang$^{1}$\thanks{Corresponding Author} \\[8pt]
$^{1}$ Singapore University of Technology and Design \\
$^{2}$ Washington University in St. Louis \\
}

\begin{document}
\maketitle
\begin{abstract}
Multicultural multi-agent systems are increasingly deployed in globally diverse settings, where different agents are grounded in different cultural backgrounds. Existing cultural evaluation focuses on value alignment: how closely a single agent matches a target culture. Yet alignment is a per-agent property and cannot reveal whether a system, taken as a whole, preserves the cultural plurality it is meant to represent. We propose value diversity as a system-level evaluation axis for multicultural agent systems, defined through the dissimilarity between culturally conditioned agents' responses on a shared value survey. Using the World Values Survey, we evaluate 19 cultures and 18 backbone models across a wide range of system configurations. We find that diversity is largely uncorrelated with alignment, indicating that the two capture complementary system properties, and that current multicultural agent systems fall substantially below human societies in value diversity. Mixed-backbone systems narrow this gap but do not close it, and the gap persists across culture compositions and agent scales. Social interaction further erodes diversity by driving agents toward consensus, and a participatory budgeting case study shows that this homogenization narrows the breadth of collective decision-making. Together, our results establish value diversity as a distinct evaluation axis for multicultural multi-agent systems and reveal a persistent homogenization tendency in current LLM-based societies. Our code and data are publicly available at \url{https://github.com/iNLP-Lab/MultiAgent-Diversity}.

\end{abstract}

\section{Introduction}
The growing capability of large language models (LLMs) has spawned a rapidly expanding ecosystem of LLM-based agents~\citep{agent1,agent2}, and multi-agent systems built from them are emerging as a major research direction~\citep{magent1,magent2}. Early work focused on improving task capability~\citep{magent3,magent4,magent5}; more recent efforts have moved toward agent-only social platforms in which no human participates~\citep{molt1, molt2, molt3, molt4, molt5}.

Existing work on the cultural dimension of LLMs has largely focused on \emph{value alignment}: an agent answers items from a social-value survey, and its responses are compared against those of a real population~\citep{align1,align2,align3,align4,align5,align6}. This paradigm evaluates how well a \emph{single agent} represents its assigned culture. However, in multicultural multi-agent systems, a property of the \emph{whole system} becomes more important: whether culturally grounded agents collectively maintain heterogeneous values when they coexist and interact~\citep{inter1,inter2,inter3,molt2,molt5}. Alignment alone is insufficient because even strongly aligned agents may still collapse toward a homogeneous value space. We therefore argue that multicultural agent systems require a distinct system-level evaluation axis, which we call \emph{value diversity}.

Defining system-level value diversity matters for two reasons. First, as a collective property of systems intended to represent or serve human societies, diversity naturally connects to the broader goal of pluralistic alignment~\citep{plural3,plural1,plural2}. Second, diversity has recently emerged as an important driver of collective decision-making and collaborative reasoning in LLM systems~\citep{magent3,reason2,agent_diverse4}. Recent work has also begun to examine diversity dynamics in multi-agent interactions~\citep{inter3,agent_diverse1,agent_diverse2,agent_diverse3}. However, the degree of diversity exhibited by an agent system remains poorly defined. Our work targets multicultural agent systems and addresses this gap from the perspective of cultural values.

Our framework extends cultural evaluation from the individual-agent level to the system level. Given a value survey and a multicultural agent system, each agent answers survey questions conditioned on its assigned cultural identity. Value alignment measures the similarity between an agent’s responses and its corresponding cultural reference, capturing local cultural fidelity. However, we define value diversity as a concept distinct from alignment. Specifically, we measure the dissimilarity between agents’ responses and aggregate these differences into a system-level diversity score, either through pairwise averaging across all agents or through structural averaging over the agent dissimilarity graph. Under this formulation, diversity is treated as a collective property of the system rather than an attribute of any individual agent.

Using the World Values Survey~\citep{wvs}, we conduct extensive experiments across 19 cultures, 18 backbone models, and a wide range of system configurations (§\ref{sec:setup}). Our main findings are as follows.

\textit{\textbf{First}, single-backbone systems are systematically less diverse than the human reference.} When a single LLM serves as the backbone for all agents in a multicultural system, none of the 18 single-backbone systems reaches the human diversity level (best system: 36.12, human: 44.07). Moreover, stronger backbone capability does not naturally translate into higher system-level value diversity (§\ref{sec:main_results}).

\textit{\textbf{Second}, diversity reveals system properties that alignment misses.} Across all systems, diversity and alignment exhibit almost no correlation, with Pearson $r=-0.12$. Several systems achieve high alignment while remaining highly homogeneous internally. Value diversity therefore complements alignment by capturing system-level properties that per-agent measures cannot reflect (§\ref{sec:alignment_vs_diversity}).

\textit{\textbf{Third}, mixed backbones improve both alignment and diversity.} In realistic deployments, different cultural agents may operate on different backbone models due to user preferences. We exhaustively explore all $18^5 \approx 1.89\text{M}$ such configurations and find that the mixed-backbone Pareto frontier consistently dominates the single-backbone frontier along both axes. Nevertheless, the gap to human-level diversity persists (§\ref{sec:heterogeneous}).

\textit{\textbf{Fourth}, neither culture selection nor agent count rescues diversity.} We vary both the cultures populating the system and the number of agents. Neither factor improves system-level diversity. More importantly, as the number of agents increases, the homogenization of multicultural agent systems becomes increasingly amplified (§\ref{sec:composition}).

\textit{\textbf{Fifth}, dynamic interaction preserves alignment but reduces diversity.} Beyond static systems, we study multi-round social exposure, where agents observe other agents' responses before generating their own. We find a more complex dynamic than predicted by Social Identity Theory~\citep{social}: although social exposure slightly improves per-agent cultural fidelity, agents drift more toward consensus, thereby reducing the system's collective plurality. Additional rounds of interaction do not recover the lost diversity (§\ref{sec:interaction}).

\textit{\textbf{Finally}}, in a democratic decision-making setting based on Participatory Budgeting, we find that high-diversity systems produce broader societal-priority coverage and greater plurality in public-resource allocation than low-diversity systems (§\ref{sec:collective}).

Together, these findings establish value diversity as a distinct and currently unmet challenge for the increasingly popular paradigm of multicultural multi-agent systems.

\section{Related Work}

\paragraph{Cultural Alignment of LLMs.} Existing cultural alignment evaluation falls into two strands. One focuses on cultural values, with most works using social-value surveys such as WVS~\citep{wvs} or Hofstede~\citep{hof} to measure how closely a model aligns with a specific culture~\citep{align1,align2,align3,align4,align5}. The other focuses on cultural knowledge, measuring how well a model reflects culture-specific commonsense and norms~\citep{know1, know2, know3, know4, know5}. Both are evaluated at the per-model level, whereas we shift the focus to the system level.

\paragraph{Multi-Agent LLM Systems.} Most multi-agent LLM systems are capability-oriented, improving task performance through agent debate and collaboration~\citep{magent3,magent4,magent5, magent6}. A second line is social-simulation-oriented, studying the dynamic behaviors that emerge among agent societies~\citep{molt1, molt2, molt3, molt4}. This direction has recently gained momentum, exemplified by Moltbook~\citep{molt5}, an agent-native social network where each agent is initialized with a distinct user identity. However, how to evaluate the cultural behavior of such systems remains an open question; we address this gap from the perspective of value diversity.

\paragraph{LLM Diversity.} Diversity has emerged as an evaluation axis distinct from accuracy or efficiency. One line focuses on human-like output diversity~\citep{diverse1, diverse2, diverse3, diverse4}. A complementary line uses diversity instrumentally as a driver of model capability~\citep{reason1, magent3, reason2} or efficient model training~\citep{train, train2}. Most recently, several works treat diversity as a key factor in multi-agent systems~\citep{agent_diverse1, agent_diverse2, agent_diverse3}. Closest to ours, \citet{inter3} studies how value diversity shapes emergent group dynamics. We instead extend the existing survey-based paradigm of value alignment into system-level value diversity.

\section{System-Level Value Diversity}
\label{sec:define}

\subsection{System Definition}

We abstract agent-driven social platforms such as MoltBook~\citep{molt5} into a \emph{multicultural agent system}. To capture cultural plurality in its minimal form, we formalize such systems as follows. We consider a multi-agent system $S = \{a_1, \ldots, a_N\}$, where each agent $a_i$ is assigned a cultural identity $c_i$ through its system prompt. The system thus contains $N$ culturally grounded agents. Prompt details in this section are provided in Appendix~\ref{sec:section3_1}.

\subsection{Answering the World Values Survey}

After system initialization, each agent $a_i$ answers value-related survey questions under its assigned cultural identity $c_i$. We instantiate this process on the World Values Survey (WVS)~\citep{wvs}, which contains $K$ value-related multiple-choice questions. For each question, the options are ordinally organized according to the degree of support for the underlying viewpoint. Each agent $a_i$ answers all questions and produces a response vector $x^{(i)} \in \mathbb{R}^K$, where each entry corresponds to the ordinal index of the selected response option.

\subsection{Value Alignment}

Prior work evaluates how closely model responses align with a target culture. Since WVS additionally provides population-level responses for each culture, we follow prior work~\citep{align3,align5} and use the majority-vote answer as the cultural ground-truth vector $\mu$, representing the prototypical value orientation of that culture. This choice is also consistent with our simplified prompt design, which aims to elicit typical cultural responses (Appendix~\ref{sec:section3_1}). The alignment of agent $a_i$ to culture $c_i$ is defined as
\begin{equation}
\label{eq:alignment}
\mathrm{Align}(x, \mu)
=
1 -
\frac{
\sqrt{\sum_{k=1}^{K}(x_k - \mu_k)^2}
}{
\sqrt{\sum_{k=1}^{K}\Delta_k^2}
},
\end{equation}
where $\Delta_k$ denotes the maximum possible disagreement on question $k$ (e.g., $\Delta_k = 3$ for a 4-point Likert item). Overall alignment is then defined as
\begin{equation}
\label{eq:system_alignment}
\mathrm{Alignment}(S)
=
\frac{1}{N}
\sum_{i=1}^{N}
\mathrm{Align}\bigl(x^{(i)}, \mu^{(i)}\bigr).
\end{equation}

\subsection{Our Metric: Value Diversity}
\label{sec:diversity}

Alignment evaluates each agent against its own cultural reference but ignores the relationships among agents within the system. To address this limitation, we introduce \emph{system-level value diversity}. Alignment and diversity capture conceptually different properties: alignment measures agent-to-human similarity, whereas diversity measures agent-to-agent dissimilarity. The two can therefore vary independently. For example, a system whose agents collapse toward culturally averaged responses may still achieve high alignment while exhibiting low diversity.

\paragraph{Pairwise Diversity}

We first define \emph{Pairwise Diversity}, which measures the average dissimilarity across all agent pairs. Given two agents with response vectors $x$ and $y$, their pairwise dissimilarity is defined as
\begin{equation}
\label{eq:diversity}
\mathrm{Div}(x, y)
=
\frac{
\sqrt{\sum_{k=1}^{K}(x_k - y_k)^2}
}{
\sqrt{\sum_{k=1}^{K}\Delta_k^2}
},
\end{equation}
and the pairwise diversity of the system is
\begin{equation}
\label{eq:pairwise_diversity}
\mathrm{Diversity}_{P}(S)
=
\frac{1}{\binom{N}{2}}
\sum_{i < j}
\mathrm{Div}\bigl(x^{(i)}, x^{(j)}\bigr).
\end{equation}

\paragraph{Structural Diversity}

$\mathrm{Diversity}_{P}(S)$ averages all $\binom{N}{2}$ pairwise distances, but many of these distances are geometrically redundant and contribute limited additional information about the system's global spread. We therefore complement $\mathrm{Diversity}_{P}(S)$ with \emph{Structural Diversity}, which averages only the $N-1$ distances required to connect all agents through the minimum spanning tree (MST) of the pairwise distance graph:
\begin{equation}
\label{eq:structural_diversity}
\mathrm{Diversity}_{S}(S)
=
\frac{1}{N-1}
\sum_{\substack{(i,j)\in\\ \mathrm{MST}(S)}}
\mathrm{Div}\bigl(x^{(i)}, x^{(j)}\bigr).
\end{equation}

Compared with $\mathrm{Diversity}_{P}(S)$, 
$\mathrm{Diversity}_{S}(S)$ provides a sharper characterization of system-level diversity by discounting redundant inter-agent relations. Pseudocode implementations of both metrics are provided in Appendix~\ref{sec:section3_2}.

For the \emph{human reference}, we replace agent responses with the corresponding majority-vote WVS response vectors for each culture and apply the same diversity metrics.

\section{Experimental Setup}
\label{sec:setup}

\paragraph{Survey} 

We use Wave 7 of the World Values Survey (WVS)~\citep{wvs}, which contains 260 value-related questions and collects human responses from 57 countries between 2017 and 2020. From the original 260 items, we retain 223 and exclude 37 questions targeting daily-life specifics rather than general value orientations.

\paragraph{Cultures}

From the 57 countries in WVS Wave 7, we select 19 spanning multiple continents: AUS, BOL, BRA, CAN, CHN, DEU, ETH, GBR, IND, KEN, MEX, NGA, NLD, NZL, RUS, THA, UKR, USA, and ZWE. Full country names and corresponding cultural identities are provided in Appendix~\ref{sec:country_details}.

\paragraph{Backbone Models}
We use 18 LLMs spanning the GPT, Claude, Gemini, Grok, Qwen, and Llama families as backbones, as shown in Table~\ref{tab:main_diversity}. We use default sampling parameters for API-based models (temperature $=1.0$, top-$p = 1.0$) and temperature $= 0.6$ for the Qwen and Llama series.

\paragraph{System Configuration}
Our main experiments use $N = 5$ agents per system, instantiated with BRA, CHN, MEX, NGA, and NZL---five cultures chosen for their substantial real-world differences. The remaining cultures are reserved for further analysis (§\ref{sec:composition}). We study two backbone settings: (i) \emph{single}, where all agents in a system share a single backbone model, and (ii) \emph{mixed}, where different agents use different backbones.

\begin{table}[t]
    \centering
    \footnotesize
    \setlength{\tabcolsep}{4pt}
    \begin{tabular}{lcc}
        \toprule
        \textbf{Model} & \textbf{Release Date} & $\mathbf{Diversity}_{P\,/\,S}$ \\
        \midrule
        gpt-5.4                       & Mar 2026 & 25.06\,/\,19.54 \\
        gpt-5-mini                    & Aug 2025 & 31.43\,/\,25.39 \\
        gpt-4o-mini                   & Jul 2024 & 25.46\,/\,19.59 \\
        \midrule
        claude-opus-4.7               & Apr 2026 & 28.44\,/\,22.36 \\
        claude-sonnet-4.5             & Sep 2025 & 26.57\,/\,21.25 \\
        claude-3.5-haiku              & Oct 2024 & 28.85\,/\,22.65 \\
        \midrule
        gemini-3.1-flash-lite$^{*}$   & Mar 2026 & 32.55\,/\,27.24 \\
        gemini-3-flash-preview        & Dec 2025 & 33.21\,/\,26.95 \\
        gemini-2.5-pro                & Jun 2025 & \textbf{36.12}\,/\,\textbf{29.60} \\
        \midrule
        grok-4.3                      & Apr 2026 & 27.42\,/\,23.93 \\
        grok-4                        & Jul 2025 & 30.60\,/\,24.66 \\
        grok-3                        & Feb 2025 & 25.18\,/\,19.82 \\
        \midrule
        Qwen3.5-27B                   & Feb 2026 & 29.82\,/\,23.25 \\
        Qwen3-32B                     & Apr 2025 & 24.61\,/\,18.97 \\
        Qwen2.5-32B-Instruct          & Sep 2024 & 20.44\,/\,16.19 \\
        \midrule
        llama-4-scout                 & Apr 2025 & 23.60\,/\,19.33 \\
        llama-3.3-70b-instruct        & Dec 2024 & 33.21\,/\,27.00 \\
        llama-3.1-70b-instruct        & Jul 2024 & 34.98\,/\,29.03 \\
        \midrule
        Human                & --       & \textbf{44.07}\,/\,\textbf{39.37} \\
        \bottomrule
    \end{tabular}
    \caption{\textbf{System-level value diversity of multicultural agent systems in the single-backbone setting.} Backbones are grouped by model family. \textbf{Bold} marks the highest LLM score per metric and the human baseline. $^{*}$\,Abbreviation of \texttt{gemini-3.1-flash-lite-preview}.}
    \label{tab:main_diversity}
\end{table}

\section{Main Diversity Results}
\label{sec:main_results}

Table~\ref{tab:main_diversity} reports $\mathrm{Diversity}_P$ and $\mathrm{Diversity}_S$ for \emph{single-backbone} systems---all five cultural agents (BRA, CHN, MEX, NGA, NZL) sharing a single backbone---alongside the human reference. All systems fall substantially below the human reference on both metrics. The most diverse system, \texttt{gemini-2.5-pro}, reaches $36.12$ / $29.60$ versus $44.07$ / $39.37$ for humans---gaps of roughly $8$ and $10$ points on a $[0,100]$ scale. The larger gap under $\mathrm{Diversity}_S$ further suggests that Structural Diversity provides a sharper characterization of system-level diversity by discounting redundant inter-agent relations, a pattern consistently observed across other backbones. Moreover, more recent backbones within a model family rarely produce the most diverse system. For example, \texttt{gpt-5.4} exhibits lower diversity than both older GPT-family backbones. This suggests that general model capability does not naturally translate into higher cultural value diversity at the system level.

\begin{figure*}[t]
    \centering
    \includegraphics[width=\textwidth]{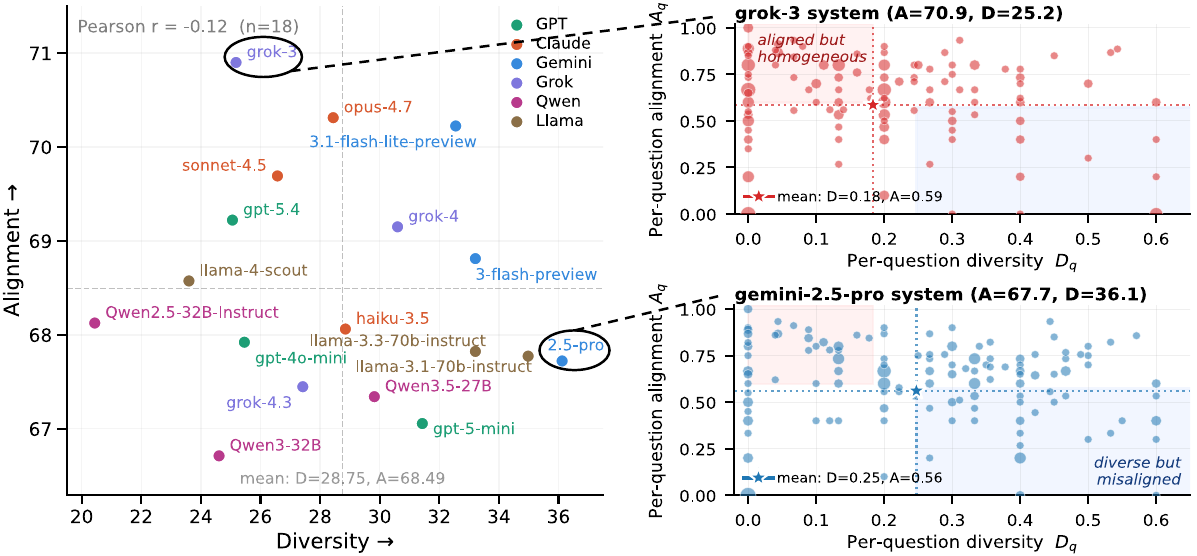}
    \caption{\textbf{Landscape of system-level value diversity and value alignment.}
    \textbf{Left:} 18 single-backbone systems on the $(Diversity, Alignment)$ plane, colored by model family. Dashed lines indicate across-system means. Pearson correlation between the two metrics is reported.
    \textbf{Right:} Per-question $(D_q, A_q)$ distributions for the two circled systems; bubble size encodes item density, and stars mark across-question means.}
    \label{fig:diversity_vs_alignment}
\end{figure*}

We also conduct a robustness check on the effect of sampling randomness. The resulting noise-diversity values range from $6.98$ to $15.21$, indicating that a non-trivial portion of the diversity reported in Table~\ref{tab:main_diversity} may be attributable to stochastic variation. Thus, the underlying model-induced diversity is likely even lower than the reported values. Detailed results are provided in Appendix~\ref{sec:noise_diversity}. Overall, these findings demonstrate the role of value diversity in revealing system-level cultural homogenization. 

In the following sections, we further examine its relationship with alignment (§\ref{sec:alignment_vs_diversity}) and investigate additional factors shaping diversity in multicultural agent systems, including mixed backbones (§\ref{sec:heterogeneous}), cultural composition, and agent count (§\ref{sec:composition}).

\subsection{Diversity Captures What Alignment Misses}
\label{sec:alignment_vs_diversity}

We define value diversity as distinct from alignment, but their relationship remains unclear. Prior work has reported tensions between alignment and diversity in synthetic LLM populations~\citep{a_d}. We examine this relationship in multicultural agent systems. We measure $\mathrm{Alignment}$ for the 18 single-backbone systems and jointly visualize it with $\mathrm{Diversity}_P$ in Figure~\ref{fig:diversity_vs_alignment} (left). Appendix~\ref{sec:section5} reports $\mathrm{Diversity}_S$ with similar findings. In the remainder of the paper, we primarily report $\mathrm{Diversity}_P$ for readability and interpretability.

Across all systems, the Pearson correlation between diversity and alignment is weak ($r=-0.12$; 95\% CI $[-0.56, 0.37]$, computed using the Fisher $z$-transformation; $n=18$), providing no clear evidence of a monotonic relationship between the two metrics. At the idealized endpoint, a perfectly human-aligned system would reproduce the human diversity level of $44.07$. However, this endpoint observation does not imply a positive relationship at intermediate levels of alignment, where systems with similar alignment may exhibit different degrees of diversity.

Importantly, the weak correlation shows that value diversity captures information not reflected by alignment alone. A system can achieve high overall alignment while still exhibiting substantial internal homogenization, as illustrated by \texttt{grok-3}. Conversely, a system with relatively high value diversity may still fail to align well with human, as represented by \texttt{gemini-2.5-pro}. We next analyze these two representative cases to better understand the relationship between diversity and alignment.

\paragraph{Two Complementary Cases}

To illustrate these complementary patterns, Figure~\ref{fig:diversity_vs_alignment} (right) decomposes the overall $(Diversity, Alignment)$ results of the \texttt{grok-3} and \texttt{gemini-2.5-pro} systems into per-question $(D_q, A_q)$ distributions. For each question $q$, $D_q$ measures the mean pairwise response distance among all agents, while $A_q$ measures the mean agent-to-human response similarity; both are normalized to $[0,1]$.

We observe that \texttt{grok-3} concentrates more heavily in the upper-left quadrant, indicating that its agents tend to remain close to the corresponding cultural ground truth (aligned) while still converging toward similar responses (homogeneous). In contrast, \texttt{gemini-2.5-pro} shifts more heavily toward the lower-right quadrant, suggesting that its agents generate more diverse but less human-aligned responses.

Taken together, these findings show that value diversity captures system properties that alignment alone cannot reflect. Knowing how well a system aligns with target cultures is insufficient to determine whether it also preserves their value plurality. This establishes value diversity as a complementary evaluation axis for multicultural agent systems.

\begin{figure}[t]
    \centering
    \includegraphics[width=\columnwidth]{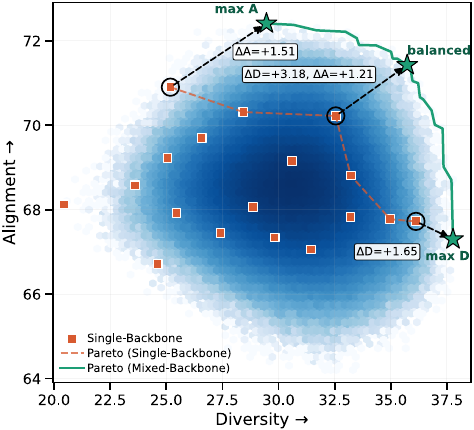}
    \caption{\textbf{Diversity--alignment landscape of all $18^5 \approx 1.89\text{M}$ backbone configurations ($N=5$ cultures).} Blue hexbin shows configuration density (darker = more). Three notable mixed-backbone configurations (green stars) are annotated with $\Delta D (Diversity), \Delta A (Alignment)$ relative to the nearest single-backbone reference (circled).}
    \label{fig:heterogeneous}
\end{figure}

\subsection{Mixed Backbones}
\label{sec:heterogeneous}
In realistic deployments, different agents may run on different backbones due to user preferences. We ask how such \emph{mixed-backbone} systems compare with their single-backbone counterparts. With $N=5$ cultural slots and $18$ candidate backbones, the configuration space contains $18^5 \approx 1.89\text{M}$ assignments, of which $18$ are single-backbone. We exhaustively evaluate all of them and visualize the resulting diversity-alignment landscape in Figure~\ref{fig:heterogeneous}.

The mixed-backbone Pareto frontier (solid green) strictly dominates the single-backbone one (dashed orange) across the entire diversity-alignment plane. Three configurations illustrate different benefits from the mixed-backbone frontier. At the alignment-optimal end, mixed backbones improve alignment by $\Delta A = +1.51$ over the best single-backbone reference; at the diversity-optimal end, they improve diversity by $\Delta D = +1.65$. Between the two extremes, a balanced configuration improves on a comparable single-backbone reference along both axes simultaneously ($\Delta D = +3.18$, $\Delta A = +1.21$). These results suggest that mixed backbones have the potential to jointly improve system-level diversity and alignment.

\begin{figure}[t]
    \centering
    \includegraphics[width=\columnwidth]{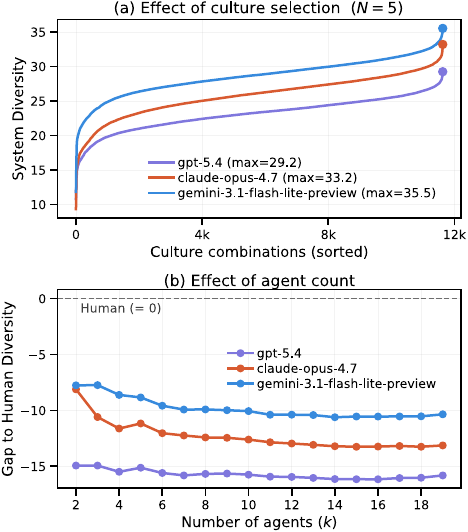}
    \caption{\textbf{Effects of culture selection and agent count on system-level value diversity.} \textbf{(a)} Diversity of all $\binom{19}{5}=11{,}628$ five-culture subsets, sorted ascending. \textbf{(b)} System-to-human diversity gap as the agent count $k$ varies. For each $k$, we report the \emph{maximum} diversity over all $\binom{19}{k}$ culture subsets for both the system and the human reference.}
    \label{fig:culture_and_count}
\end{figure}

\subsection{Effects of Cultural Composition and Agent Count}
\label{sec:composition}
Two additional factors shape a multicultural agent system: which cultures populate it, and how many agents it contains. Here, we investigate how these two factors affect system-level diversity. We select the three most advanced backbones: \texttt{gpt-5.4}, \texttt{claude-opus-4.7}, and \texttt{gemini-3.1-flash-lite-preview}, and fix each as the single backbone of a system. For each, we (i) fix $N=5$ and exhaustively evaluate all systems built from $\binom{19}{5} = 11{,}628$ culture subsets, and (ii) vary the agent count $k$ from $2$ to $19$, reporting the maximum diversity over all $\binom{19}{k}$ subsets at each $k$.

\paragraph{Culture Composition (Figure~\ref{fig:culture_and_count}a)}
Even the highest-diversity subset reaches only $29.2$--$35.5$ across the three backbones. Most configurations cluster within the middle range ($20$--$30$), indicating that nearly any five-culture sample yields similar moderate diversity.

\paragraph{Agent Count (Figure~\ref{fig:culture_and_count}b)}
To account for the natural reduction in average pairwise distance as the number of agents increases, we report the relative difference from the human reference for each agent count $k$.\footnote{Inter-agent distances may shrink as $k$ grows simply because the response space becomes more densely populated; comparing against the human reference at the same $k$ controls for this baseline effect.} As shown, although the difference is relatively small for small $k$, the disparity between LLM-based systems and human societies becomes increasingly pronounced as $k$ grows.

Overall, these results suggest that culture selection alone provides limited gains in system-level value diversity. As the number of agents increases, the initial homogenization of multicultural agent systems becomes increasingly amplified.

\begin{figure}[t]
    \centering
    \includegraphics[width=\columnwidth]{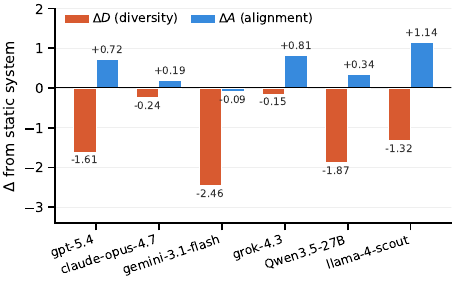}
    \caption{\textbf{Effect of one-round social exposure on system-level diversity and alignment.} Each value is the change relative to the static system of Section~\ref{sec:main_results}. All systems lose diversity ($\Delta D < 0$), while alignment generally rises but by a much smaller margin.}
    \label{fig:social_exposure}
\end{figure}

\section{Towards Dynamic Interaction}
\label{sec:interaction}
Our experiments so far focus on static systems, where each agent answers the WVS independently. Real-world agent-driven platforms, however, are inherently interactive. We ground our interaction experiments in Social Identity Theory~\citep{social}, which suggests that exposure to out-group members can strengthen in-group identification and sharpen between-group distinctiveness. We thus ask whether dynamic interaction increases system-level diversity by reinforcing each agent's culturally distinctive positions.

\subsection{Experimental Setup}

We study dynamic interaction through \emph{multi-round social exposure}. In each round, before answering a WVS item, each agent is shown the previous-round responses of the other $N-1$ agents to the same item and then produces its own answer. We iterate this process for $K$ rounds, where round $0$ corresponds to the static system in Section~\ref{sec:main_results}. We report system-level value diversity and alignment after each round of exposure and compare the results against the static baseline. We adopt the same configuration as the main experiments: five cultural agents ($N=5$) with cultures BRA, CHN, MEX, NGA, and NZL sharing a single backbone model. Prompt details are provided in Appendix~\ref{sec:section6_1}.

\subsection{Results}

\paragraph{Diversity decreases under social exposure.}

We conduct experiments on representative backbones across six model families. Figure~\ref{fig:social_exposure} shows the results after one round of social exposure. Every system loses diversity under social exposure, with an average decrease of $\Delta D = -1.27$. Alignment, in contrast, generally increases by a much smaller margin. These findings suggest a more complex dynamic than predicted by Social Identity Theory. Although social exposure slightly improves per-agent cultural fidelity, agents do not respond by reinforcing culturally distinctive positions. Instead, agents tend to drift toward consensus, thereby reducing the system's collective plurality.

\begin{figure}[t]
    \centering
    \includegraphics[width=\columnwidth]{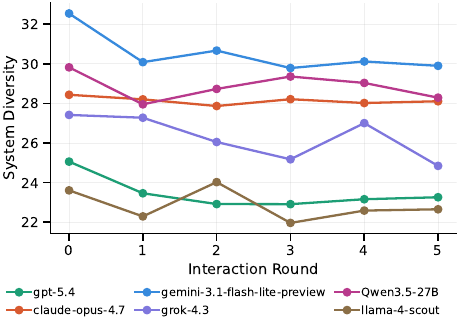}
    \caption{\textbf{System diversity over five rounds of interaction} for six representative systems.}
    \label{fig:multiround_diversity}
\end{figure}

\begin{figure*}[t]
\centering
\includegraphics[width=0.98\textwidth]{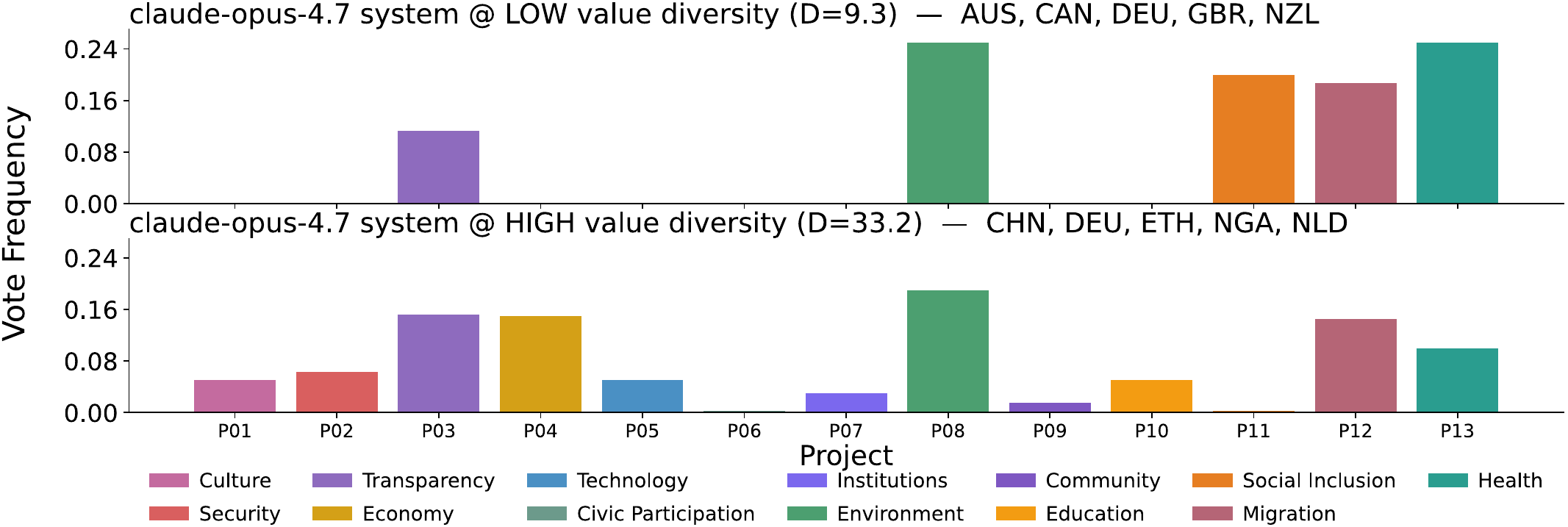}
\caption{
\textbf{Collective decision-making outcomes in the Participatory Budgeting task.}
Both systems use \texttt{claude-opus-4.7} as the backbone model.
The results show that the high-diversity system distributes support across substantially broader societal dimensions.
}
\label{fig:wvs_pb_case}
\end{figure*}

\paragraph{Multi-round interaction does not recover diversity.}

Figure~\ref{fig:multiround_diversity} shows the evolution of system diversity under multi-round interaction.\footnote{The corresponding alignment dynamics are reported in Appendix~\ref{sec:section6_2}.} After $K>1$ rounds of exposure, some systems, such as \texttt{claude-opus-4.7}, maintain the diversity level reached after the first interaction round. Many other systems, however, continue to lose diversity, albeit with moderate fluctuations. Crucially, no system recovers toward its round-$0$ diversity baseline. These results suggest that homogenization induced by social exposure represents a lasting tendency of interactive multicultural agent systems, rather than merely a transient effect.

Overall, these findings suggest that the static diversity measured in the main experiments may overestimate the diversity these systems would exhibit in realistic interactive deployments.

\section{Case Study: Diversity in Collective Decision-Making}
\label{sec:collective}

We next present a case study examining whether systems with low and high value diversity exhibit different patterns of collective decision-making.

\subsection{Experimental Setup}

\paragraph{Participatory Budgeting} To study this question, we consider a democratic decision-making scenario based on Participatory Budgeting (PB), where citizens directly propose and vote on urban projects to determine how limited public resources should be allocated~\citep{pb}. Rather than using projects from a specific municipality, we directly construct the candidate pool using the 13 value dimensions in the WVS dataset, each corresponding to a socially beneficial public project. The project categories cover Culture, Security, Transparency, Economy, Technology, Civic Participation, Institutions, Environment, Community, Education, Social Inclusion, Migration, and Health.

\paragraph{Decision-Making}
After initializing a system consisting of $N$ culturally grounded agents, each agent independently votes for 4 projects out of the 13 candidates. To obtain statistically stable behavioral patterns, we sample each agent 20 times, resulting in $N \times 4 \times 20$ project approvals in total. We then compute the aggregated vote-frequency distribution over all projects. By comparing high- and low-diversity systems, we analyze how value diversity shapes collective societal prioritization.

\paragraph{Agent System} We use \texttt{claude-opus-4.7} as the backbone. The number of agents in a system is fixed to $N=5$. Based on the exhaustive cultural-composition analysis in Section~\ref{sec:composition}, we select the lowest-diversity and highest-diversity systems. We conduct the participatory budgeting task independently for each system, resulting in vote-frequency distributions over $400$ project approvals per system.

We conduct the same experiment using \texttt{gpt-5.4} and \texttt{gemini-3.1-flash-lite-preview}, both of which exhibit trends consistent with the \texttt{claude-opus-4.7} results. Full project descriptions, prompts, and additional backbone results are provided in the appendix~\ref{sec:section7}.

\subsection{Results}

Figure~\ref{fig:wvs_pb_case} compares the aggregated vote-frequency distributions produced by low- and high-diversity systems. The low-diversity system exhibits highly concentrated collective decision-making behavior, with approvals collapsing onto only a few societal dimensions. In contrast, the high-diversity system produces substantially broader societal coverage.

This case study illustrates that systems with different levels of value diversity can exhibit markedly different downstream collective behaviors. In this setting, the more diverse system shows broader societal-priority coverage and greater plurality in public-resource allocation. We note, however, that because cultural composition differs between the two systems, these results are descriptive rather than a causal isolation of the effect of diversity.

\section{Conclusion}

We propose \emph{value diversity} as a system-level evaluation axis for multicultural multi-agent systems, complementing the per-agent \emph{value alignment} paradigm. Across 19 cultures, 18 backbones, and millions of system configurations, we find empirically that diversity complements alignment as a distinct evaluation axis. All single-backbone systems fall below the human diversity reference, and although mixed backbones narrow this gap, substantial differences from human societies persist across culture selection and agent scaling. Social interaction further erodes diversity by driving agents toward consensus, and a participatory budgeting case study shows that this homogenization directly narrows collective decision-making. Together, our results establish value diversity as a distinct and currently unmet challenge for increasingly realistic LLM-based societies.

\section*{Limitations}

\paragraph{Simplified Interaction and Decision-Making Settings}
Our interaction and collective decision-making experiments adopt simplified settings, including social exposure and WVS-grounded participatory budgeting. The low- and high-diversity configurations in the participatory budgeting study additionally differ in cultural composition as well as measured diversity, which our current design does not fully disentangle. More realistic agent interactions, social networks, and real-world policy environments remain important directions for future work.

\paragraph{Cultural Prototype as Reference}
Our framework compares culturally conditioned agent responses against WVS population majority votes, treating both as prototypical value orientations. While our prompts explicitly elicit typical cultural responses, a single LLM sample remains an imperfect approximation of the underlying cultural prototype, whereas WVS majority votes aggregate over large human populations. This asymmetry may affect the absolute comparability between human and model diversity scores. Moreover, WVS-7 responses were collected between 2017 and 2020, and our scores should be interpreted relative to this temporal snapshot rather than a timeless cultural reference.

\paragraph{Generalizability}
Our framework adopts a deliberately minimal abstraction of multicultural multi-agent systems, with the goal of remaining applicable to more complex settings. Whether the homogenization patterns reported here generalize to richer cultural signals---such as everyday dialogue, normative reasoning, and emergent behaviors on agent-native platforms---remains an open question, and we leave instantiations of our framework in such settings to future work.

\section*{Acknowledgements}
This research/project is supported by the National Research Foundation, Singapore under its NRF Fellowship (Award No: NRF-NRFF18-2026-0012) and by the National Research Foundation Singapore under the AI Singapore Programme (AISG Award No: AISG3-RPGV-2025-016).

\bibliography{custom}
\clearpage

\appendix

\section{System-Level Value Diversity Details}
\label{sec:section3}

\subsection{Agent Prompt}
\label{sec:section3_1}

After system initialization, each agent answers WVS questions under its assigned cultural identity. For an agent $a_i$ with culture $c_i$, we use the following system prompt:

\begin{quote}
\small
\texttt{
You are a respondent from \{country\}. Answer questions based on typical cultural values in \{culture\} culture.
}
\end{quote}

For each WVS question, the user prompt is:

\begin{quote}
\small
Question: \texttt{\{question\}}

\vspace{2pt}

Output format MUST be exactly:
\texttt{\textbackslash boxed\{number\}}
\end{quote}

\paragraph{Discussion} Importantly, the prompt is intentionally minimal. We do not provide detailed demographic profiles, behavioral descriptions, or culture-specific stereotypes beyond the target cultural identity itself. This design follows prior survey-based cultural alignment work~\citep{align3,align5} and aims only to activate the model's internal representation of the corresponding culture. As a result, the measured alignment and diversity primarily reflect the backbone model's intrinsic cultural knowledge and value associations, rather than effects introduced by complex prompt engineering.

\subsection{Implementation of Diversity Metrics}
\label{sec:section3_2}

Algorithm~\ref{alg:pairwise_diversity} and Algorithm~\ref{alg:structural_diversity} summarize the implementations of Pairwise Diversity and Structural Diversity used throughout the paper.

Structural Diversity is implemented through minimum spanning tree (MST) extraction using the SciPy graph library\footnote{\url{https://docs.scipy.org/doc/scipy/reference/generated/scipy.sparse.csgraph.minimum_spanning_tree.html}}.

\section{Country Details in Experiments}
\label{sec:country_details}

Table~\ref{tab:culture_details} lists the country codes, full country names, and corresponding cultural identities used throughout the experiments.

\begin{table}[h]
\centering
\small
\setlength{\tabcolsep}{10pt}
\begin{tabular}{ccc}
\toprule
\textbf{Code} & \textbf{Country} & \textbf{Cultural Identity} \\
\midrule
AUS & Australia & Australian \\
BOL & Bolivia & Bolivian \\
BRA & Brazil & Brazilian \\
CAN & Canada & Canadian \\
CHN & China & Chinese \\
DEU & Germany & German \\
ETH & Ethiopia & Ethiopian \\
GBR & United Kingdom & British \\
IND & India & Indian \\
KEN & Kenya & Kenyan \\
MEX & Mexico & Mexican \\
NGA & Nigeria & Nigerian \\
NLD & Netherlands & Dutch \\
NZL & New Zealand & New Zealander \\
RUS & Russia & Russian \\
THA & Thailand & Thai \\
UKR & Ukraine & Ukrainian \\
USA & United States & American \\
ZWE & Zimbabwe & Zimbabwean \\
\bottomrule
\end{tabular}
\caption{Countries and cultural identities used in experiments.}
\label{tab:culture_details}
\end{table}

\begin{table}[t]
    \centering
    \footnotesize
    \setlength{\tabcolsep}{5pt}
    \begin{tabular}{lcc}
        \toprule
        \textbf{Model} & $\mathbf{Diversity}_P$ & $\mathbf{Diversity}_S$ \\
        \midrule
        gpt-5.4                         & 8.25  & 6.98 \\
        gpt-5-mini                      & 12.49 & 11.82 \\
        gpt-4o-mini                     & 7.70  & 7.15 \\
        claude-opus-4.7                 & 4.57  & 3.83 \\
        claude-sonnet-4.5               & 8.63  & 8.29 \\
        gemini-3.1-flash-lite$^{*}$   & 8.63  & 7.68 \\
        gemini-3-flash-preview          & 11.64 & 10.55 \\
        gemini-2.5-pro                  & 12.43 & 11.21 \\
        grok-4.3                        & 15.21 & 14.28 \\
        \bottomrule
    \end{tabular}
    \caption{Noise diversity when all five agents are assigned to \textsc{CHN} and use the same backbone. Thus, the measured diversity provides a reference for variation induced by stochastic sampling rather than differences in cultural assignments. $^{*}$\,Abbreviation of \texttt{gemini-3.1-flash-lite-preview}.}
    \label{tab:noise_diversity}
\end{table}

\begin{algorithm}[t]
\small
\caption{Pairwise Diversity Implementation}
\label{alg:pairwise_diversity}

\begin{algorithmic}[1]

\Require Agent response dictionary $\texttt{all\_answers}$
\Require WVS question metadata $\texttt{ref\_dict}$

\State $\texttt{agents} \leftarrow$ sorted agent list
\State $\texttt{pair\_dict} \leftarrow \{\}$
\State $\texttt{pairwise\_scores} \leftarrow [\ ]$

\For{each agent pair $(a_i, a_j)$}

    \State $\texttt{vec1} \leftarrow \texttt{all\_answers}[a_i]$
    \State $\texttt{vec2} \leftarrow \texttt{all\_answers}[a_j]$

    \State $\texttt{common\_qs} \leftarrow$
    shared answered questions

    \State $\texttt{squared\_sum} \leftarrow 0$
    \State $\texttt{max\_squared\_sum} \leftarrow 0$

    \For{each question $q \in \texttt{common\_qs}$}

        \If{$q \notin \texttt{ref\_dict}$}
            \State continue
        \EndIf

        \State $\texttt{delta} \leftarrow$
        number of options for $q$ minus $1$

        \If{$\texttt{delta}=0$}
            \State continue
        \EndIf

        \State Accumulate:
        \[
        \texttt{squared\_sum}
        \mathrel{+}= (x_q^{(i)} - x_q^{(j)})^2
        \]

        \State Accumulate:
        \[
        \texttt{max\_squared\_sum}
        \mathrel{+}= \texttt{delta}^2
        \]

    \EndFor

    \If{$\texttt{max\_squared\_sum}=0$}
        \State continue
    \EndIf

    \State Compute normalized distance:
    \[
    d_{ij}
    =
    \frac{
    \sqrt{\texttt{squared\_sum}}
    }{
    \sqrt{\texttt{max\_squared\_sum}}
    }
    \]

    \State Append $d_{ij}$ to $\texttt{pairwise\_scores}$
    \State Store $d_{ij}$ in $\texttt{pair\_dict}$

\EndFor

\State Return:
\[
\mathrm{Diversity}_{P}(S)
=
\mathrm{mean}(\texttt{pairwise\_scores})
\]

\end{algorithmic}
\end{algorithm}

\begin{algorithm}[t]
\small
\caption{Structural Diversity Implementation}
\label{alg:structural_diversity}

\begin{algorithmic}[1]

\Require Pairwise distance dictionary $\texttt{pair\_dict}$
\Require Agent list $\texttt{agents}$

\State $N \leftarrow |\texttt{agents}|$

\If{$N < 2$ or $\texttt{pair\_dict}$ is empty}
    \State return $0$
\EndIf

\State Initialize symmetric distance matrix:
\[
M \in \mathbb{R}^{N \times N}
\]

\For{each agent pair $(a_i, a_j)$}

    \State Retrieve pairwise distance:
    \[
    d_{ij}
    \leftarrow
    \texttt{pair\_dict}(a_i,a_j)
    \]

    \State Fill:
    \[
    M[i,j] \leftarrow d_{ij}
    \]
    \[
    M[j,i] \leftarrow d_{ij}
    \]

\EndFor

\State Compute minimum spanning tree:
\[
T \leftarrow \mathrm{MST}(M)
\]

\State Compute:
\[
\texttt{mst\_span}
=
\frac{
\sum_{(i,j)\in T} M[i,j]
}{
N-1
}
\]

\State Return:
\[
\mathrm{Diversity}_{S}(S)
=
\texttt{mst\_span}
\]

\end{algorithmic}
\end{algorithm}

\section{Robustness Check}
\label{sec:noise_diversity}

Our diversity metrics are computed from sampled agent responses and may therefore capture both meaningful between-agent variation and sampling randomness. To estimate this noise floor, we assign all five agents to \textsc{CHN} while using the same backbone for every agent. Because the agents share the same cultural assignment, any measured diversity cannot be attributed to cross-cultural differences; instead, it provides a reference for variation arising from stochastic response sampling and the measurement pipeline.

Table~\ref{tab:noise_diversity} reports the resulting noise-diversity scores for nine representative backbones. The values range from $4.57$ to $15.21$ for $\mathrm{Diversity}_P$ and from $3.83$ to $14.28$ for $\mathrm{Diversity}_S$, indicating that sampling randomness can contribute non-trivially to the measured diversity.

This analysis serves as a reference rather than an exact correction, since sampling noise may not be linearly separable from culturally induced variation. Nevertheless, the non-negligible noise-floor values suggest that the underlying model-induced diversity is likely lower than the raw scores reported in Table~\ref{tab:main_diversity}. This further reinforces our main conclusion that current LLM-based multicultural agent systems remain substantially less value-diverse than the human reference.

\section{Additional Results on Diversity--Alignment Relationship}
\label{sec:section5}

\begin{figure}[t]
    \centering
    \includegraphics[width=\columnwidth]{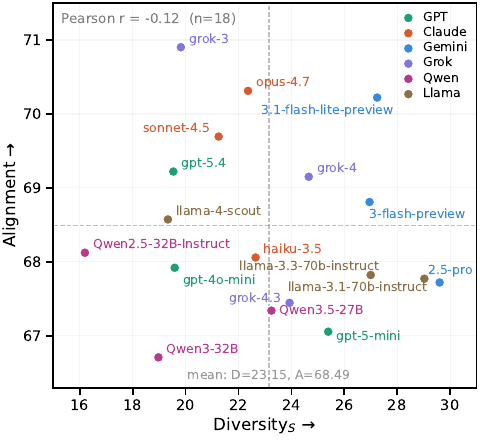}
    \caption{\textbf{Relationship between Structural Diversity and Alignment.}
    Each point represents one single-backbone multicultural agent system. Dashed lines indicate across-system means.}
    \label{fig:diversity_vs_alignment_structural}
\end{figure}

Figure~\ref{fig:diversity_vs_alignment_structural} reports the relationship between $\mathrm{Alignment}$ and $\mathrm{Diversity}_S$ (Structural Diversity). Similar to the pairwise metric in the main paper, the correlation between the two remains weak, further supporting the conclusion that alignment does not reliably reflect whether a system preserves internal cultural plurality.

Compared with $\mathrm{Diversity}_P$, $\mathrm{Diversity}_S$ produces a sharper separation between systems with genuine structural plurality and those whose diversity mainly arises from redundant pairwise disagreement. In particular, several highly aligned systems exhibit noticeably lower structural diversity, suggesting that their agents converge toward structurally homogeneous value configurations despite maintaining reasonable alignment with cultural references.

\section{Dynamic Interaction Details}
\label{sec:section6}

\subsection{Prompt}
\label{sec:section6_1}

\paragraph{System Prompt}

Each agent is initialized with the same culturally grounded system prompt used in the main experiments:

\begin{quote}
\small
\texttt{
You are a respondent from \{country\}. Answer questions based on typical cultural values in \{culture\} culture.
}
\end{quote}

\paragraph{Social Exposure Prompt}

During dynamic interaction, before answering each WVS item, every agent is additionally shown the responses generated by the other agents in the previous interaction round. The user instruction is constructed as:

\begin{quote}
\small
\texttt{
Question: \{WVS question\}
}
\vspace{0.3em}

\texttt{
Here are answers from people of other cultures:
}
\vspace{0.3em}

\texttt{
\{culture\_1\}: \{response\_1\}
}
\vspace{0.2em}

\texttt{
\{culture\_2\}: \{response\_2\}
}
\vspace{0.2em}

\texttt{
...
}
\vspace{0.3em}

\texttt{
You may consider these answers before making your decision.
}
\vspace{0.3em}

\texttt{
Output format MUST be exactly:
}
\vspace{0.3em}

\texttt{
\textbackslash boxed\{number\}
}
\end{quote}

\begin{figure}[t]
    \centering
    \includegraphics[width=\linewidth]{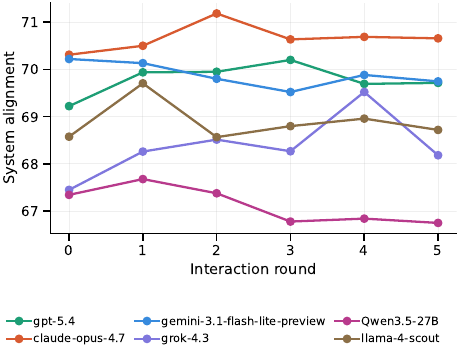}
    \caption{System alignment over five rounds of interaction for the six representative systems.}
    \label{fig:multiround_alignment}
\end{figure}

\subsection{Multi-turn Alignment Dynamics}
\label{sec:section6_2}
Figure~\ref{fig:multiround_alignment} shows the corresponding alignment dynamics for the same six representative systems.

In contrast to the consistent diversity loss observed in the main paper, alignment exhibits no systematic trend across rounds. Most systems fluctuate within a narrow band around their round-$0$ value, and no system shows a sustained gain or loss comparable in magnitude to the diversity drop. This asymmetry reinforces our main finding: social exposure primarily reshapes the system's collective plurality rather than its per-agent cultural fidelity. In other words, homogenization under interaction is a system-level phenomenon that alignment-based evaluation alone would fail to detect.

\section{Collective Decision-Making Details}
\label{sec:section7}

\paragraph{Project Descriptions}

Table~\ref{tab:wvs_pb_projects} lists the 13 projects used in the participatory budgeting task. Each project is derived from a corresponding value dimension in the WVS dataset. The project names and descriptions were generated using GPT-5.5 to transform abstract value dimensions into concrete public-policy initiatives suitable for participatory budgeting scenarios.

\begin{table*}[t]
\centering
\footnotesize
\renewcommand{\arraystretch}{1.15}
\setlength{\tabcolsep}{4pt}

\begin{tabular}{c p{3.4cm} p{2.0cm} p{2.5cm} p{6.35cm}}
\toprule
ID & Project Name & Category & WVS Dimension & Description \\
\midrule

01 &
Traditional Religious Heritage Festival &
Culture &
Religious Values &
Funding community festivals and educational programs that preserve traditional religious customs, ceremonies, and local cultural heritage. \\

02 &
Smart Public Safety Camera Network &
Security &
Security &
Expanding AI-assisted public surveillance cameras and emergency response systems to improve urban safety and crime prevention. \\

03 &
Transparent Public Spending Platform &
Transparency &
Corruption &
Creating an open digital platform that allows citizens to track government spending, procurement, and public project budgets. \\

04 &
Industrial Growth and Job Creation Program &
Economy &
Economic Values &
Supporting local manufacturing, industrial infrastructure, and employment growth through public investment programs. \\

05 &
Public AI and Technology Innovation Hub &
Technology &
Science \& Technology &
Establishing community AI labs, robotics education centers, and startup incubation programs for emerging technologies. \\

06 &
Community Civic Participation Centers &
Civic Participation &
Political Interest \& Political Participation &
Funding local civic centers that organize public discussions, citizen assemblies, and participatory decision-making activities. \\

07 &
National Institutions and Public Leadership Education &
Institutions &
Political Culture \& Political Regimes &
Supporting public education programs on national institutions, governance systems, civic responsibility, and political leadership. \\

08 &
Climate Sustainability and Green Energy Initiative &
Environment &
Postmaterialist Index &
Investing in renewable energy, carbon reduction projects, urban sustainability programs, and environmental protection initiatives. \\

09 &
Neighborhood Community Cooperation Program &
Community &
Social Capital, Trust \& Organizational Membership &
Supporting local volunteer organizations, neighborhood associations, and public community-building activities. \\

10 &
Public Ethics and Integrity Education Campaign &
Education &
Ethical Values and Norms &
Funding educational campaigns promoting honesty, social responsibility, anti-corruption awareness, and ethical public behavior. \\

11 &
Social Inclusion and Diversity Support Initiative &
Social Inclusion &
Social Values, Attitudes \& Stereotypes &
Supporting programs that promote social inclusion, reduce discrimination, and encourage understanding across different social groups. \\

12 &
Immigrant Integration and Language Support Centers &
Migration &
Migration &
Providing language education, job assistance, and community integration services for immigrants and refugees. \\

13 &
Public Mental Health and Well-being Initiative &
Health &
Happiness and Well-being &
Expanding public mental health support, community wellness activities, and accessible counseling services. \\

\bottomrule
\end{tabular}

\caption{
Public projects used in the WVS-grounded participatory budgeting task.
Each project is derived from one WVS value dimension.
}

\label{tab:wvs_pb_projects}

\end{table*}

\paragraph{Prompt}

The culturally grounded system prompt is constructed as:

\begin{quote}
\small
\texttt{
You are a respondent from \{country\}. Answer based on typical societal priorities, cultural values, and public policy preferences commonly associated with \{culture\_name\} culture.
}
\end{quote}

The participatory budgeting instruction prompt is:

\begin{quote}
\small
\texttt{
A national participatory budgeting program is allocating limited public funding across societal development initiatives.
}
\vspace{0.3em}

\texttt{
There are 13 candidate projects:
}
\vspace{0.3em}

\texttt{
[Project list (ID, project name, category, WVS dimension, and description) omitted for brevity]
}
\vspace{0.3em}

\texttt{
Voting instruction:
}
\begin{itemize}
    \item \texttt{Select EXACTLY 4 projects that should receive funding.}
    \item \texttt{Base your choices on long-term societal priorities and values.}
    \item \texttt{Consider cultural preferences, governance priorities, social development, and public well-being.}
\end{itemize}

\texttt{
Output format MUST be exactly:
}
\vspace{0.3em}

\texttt{
\textbackslash boxed\{id1, id2, id3, id4\}
}
\vspace{0.3em}

\texttt{
For example:
}
\vspace{0.3em}

\texttt{
\textbackslash boxed\{2, 5, 8, 11\}
}
\end{quote}

\begin{figure*}[t]
\centering
\includegraphics[width=0.98\textwidth]{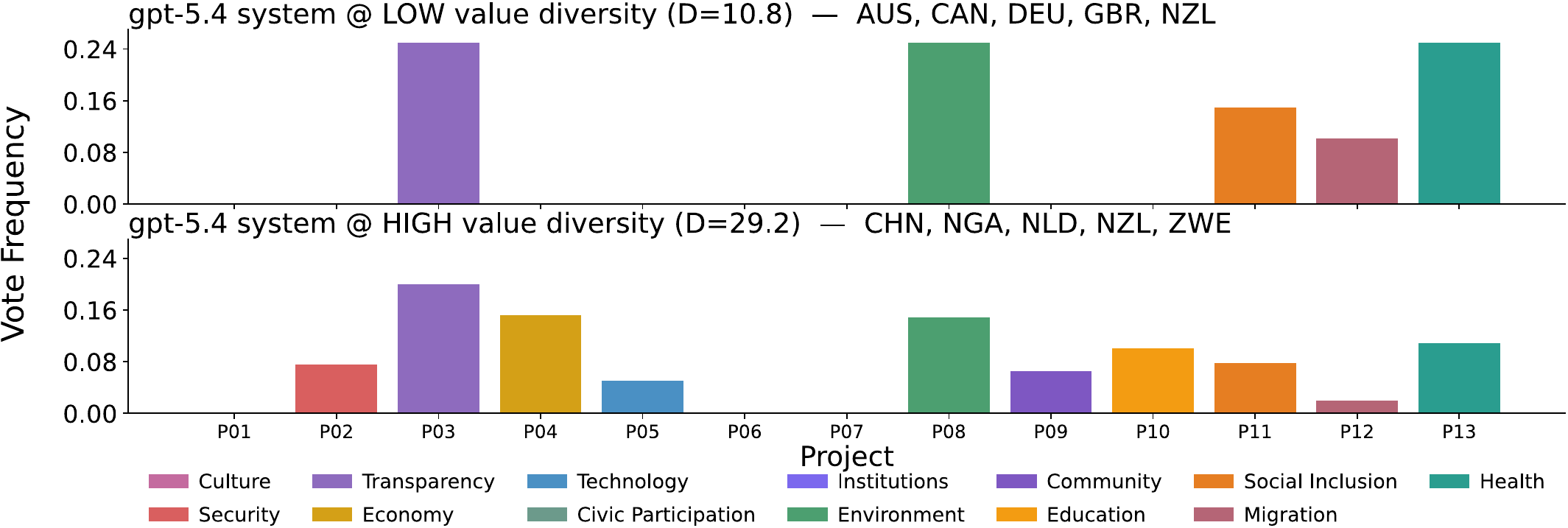}
\caption{
Collective decision-making outcomes in the Participatory Budgeting task.
Both systems use \texttt{gpt-5.4} as the backbone model.
The results show that the high-diversity system distributes support across substantially broader societal dimensions.
}
\label{fig:wvs_pb_case_gpt}
\end{figure*}

\begin{figure*}[t]
\centering
\includegraphics[width=0.98\textwidth]{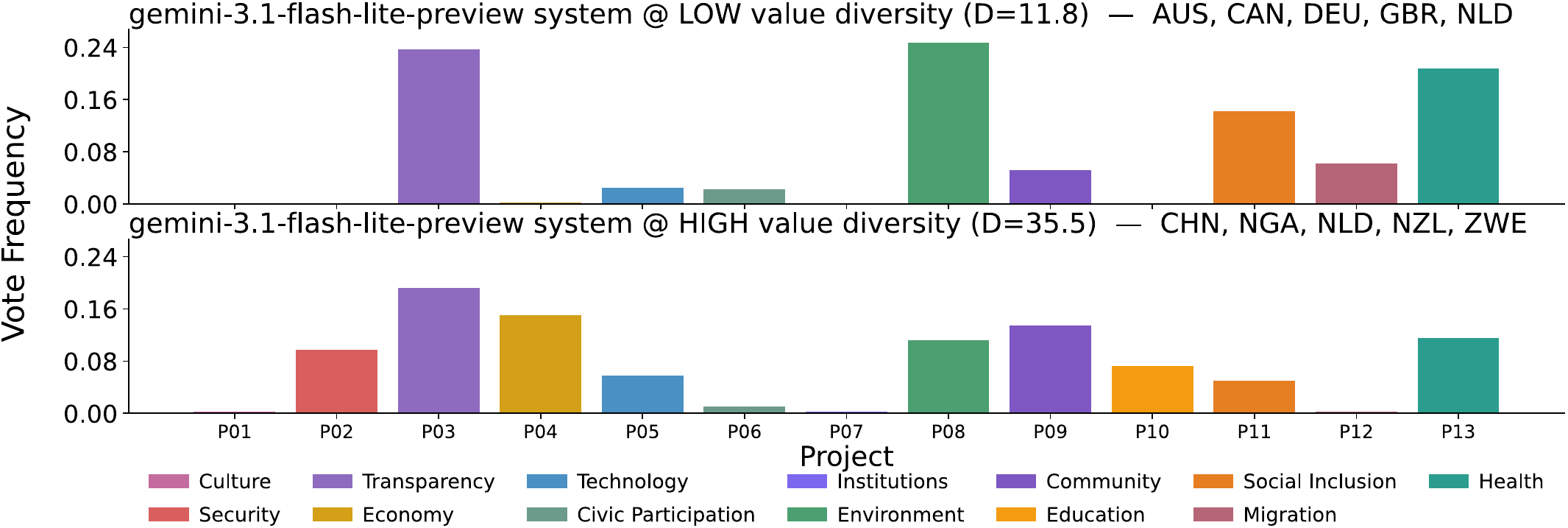}
\caption{
Collective decision-making outcomes in the Participatory Budgeting task.
Both systems use \texttt{gemini-3.1-flash-lite-preview} as the backbone model.
The results show that the high-diversity system distributes support across substantially broader societal dimensions.
}
\label{fig:wvs_pb_case_gemini}
\end{figure*}

\paragraph{Additional Results}

Additional results using \texttt{gpt-5.4} and \texttt{gemini-3.1-flash-lite-preview} are shown in Figure~\ref{fig:wvs_pb_case_gpt} and Figure~\ref{fig:wvs_pb_case_gemini}, respectively. Both backbones exhibit trends consistent with the \texttt{claude-opus-4.7} results presented in the main paper.

\end{document}